\documentclass{article} 
\usepackage[dvipsnames,table]{xcolor}

\usepackage{tccml_iclr2025_conference,times}

\usepackage{amsmath,amsfonts,bm}

\def\eqref#1{equation~\ref{#1}}

\def\1{\bm{1}}

\DeclareMathAlphabet{\mathsfit}{\encodingdefault}{\sfdefault}{m}{sl}
\SetMathAlphabet{\mathsfit}{bold}{\encodingdefault}{\sfdefault}{bx}{n}

\usepackage{hyperref}
\usepackage{url}
\usepackage{graphicx}

\title{WeatherMesh-3: Fast and accurate operational global weather forecasting}

\author{Haoxing Du\thanks{Equal contribution, corresponding authors} \And Lyna Kim\footnotemark[1] \And Joan Creus-Costa \And Jack Michaels \And Anuj Shetty \And Todd Hutchinson \And Christopher Riedel \And John Dean \And \\
WindBorne Systems\\
Palo Alto, CA 94303, USA \\
\texttt{\{haoxing, lyna\}@windbornesystems.com} \\
}

\definecolor{nred}{rgb}{0.9,0.1,0.1}
\definecolor{nblack}{rgb}{0,0,0}
\definecolor{nblue}{rgb}{0.2,0.2,0.8}
\definecolor{ngreen}{rgb}{0.2,0.6,0.2}

\iclrfinalcopy 
\begin{document}

\maketitle

\begin{abstract}
\hyphenpenalty=10000
\exhyphenpenalty=10000
We present WeatherMesh-3 (WM-3), an operational transformer-based global weather forecasting system that improves the state of the art in both accuracy and computational efficiency. We introduce the following advances: 1) a latent rollout that enables arbitrary-length predictions in latent space without intermediate encoding or decoding; and 2) a modular architecture that flexibly utilizes mixed-horizon processors and encodes multiple real-time analyses to create blended initial conditions. WM-3 generates 14-day global forecasts at 0.25-degree resolution in 12 seconds on a single RTX 4090. This represents a $>$100,000-fold speedup over traditional NWP approaches while achieving superior accuracy with up to 37.7\% improvement in RMSE over operational models, requiring only a single consumer-grade GPU for deployment. We aim for WM-3 to democratize weather forecasting by providing an accessible, lightweight model for operational use while pushing the performance boundaries of machine learning-based weather prediction. 
\end{abstract}

\section{Introduction}\label{sec:intro}

Numerical Weather Prediction (NWP) systems are fundamental to modern society, 
providing critical guidance for public safety, economic planning, and climate adaptation. However, operating NWP systems poses a high computational cost, particularly for developing regions and smaller organizations seeking to deliver essential real-time forecasting products. 

Recent advances in machine learning-based weather prediction (MLWP) have demonstrated the potential to alter this paradigm~\citep{kurth2023fourcastnet, chen2023fuxi, price2023gencast, lang2024aifs}. Models such as GraphCast or Aurora have shown that neural architectures can match or exceed the accuracy of traditional NWP systems while reducing computational requirements by several orders of magnitude~\citep{lam2023graphcast, bodnar2024aurora}. 

However, current AI approaches still face several limitations for operational use. First, despite promising results, MLWP models must continue to demonstrate improved forecast skill relative to operational NWP systems in real-time settings. Second, these models often produce excessively blurred forecasts, a direct consequence of architectural and training decisions that introduce accumulating error and can limit operational utility. Third, while less computationally demanding than traditional NWP, MLWP systems still require substantial computational resources and expertise to operate, creating barriers to widespread adoption.

We present WeatherMesh-3 (WM-3), an operational transformer-based global forecasting system that introduces advances to directly mitigate these limitations. WM-3 dramatically reduces both hardware requirements and operational complexity while improving state-of-the-art MLWP performance. 
\section{Methods}\label{sec:methods}

\subsection{Model architecture}

WeatherMesh-3 takes a single state of global weather as input to predict the state of global weather any integer number of hours later.
It represents the weather state on a latitude-longitude grid at 0.25 degree resolution, with surface and atmospheric variables at each grid point.
Like Pangu-Weather~\citep{bi2023pangu}, WM-3 makes heavy use of the vision transformer architecture~\citep{dosovitskiy2020vit}, and treats pressure level as a third dimension.
See Appendix~\ref{app:arch} for details on the model architecture.

\paragraph{Encoder-processor-decoder architecture}

\begin{figure}[h!]
    \centering
    \includegraphics[width=\linewidth]{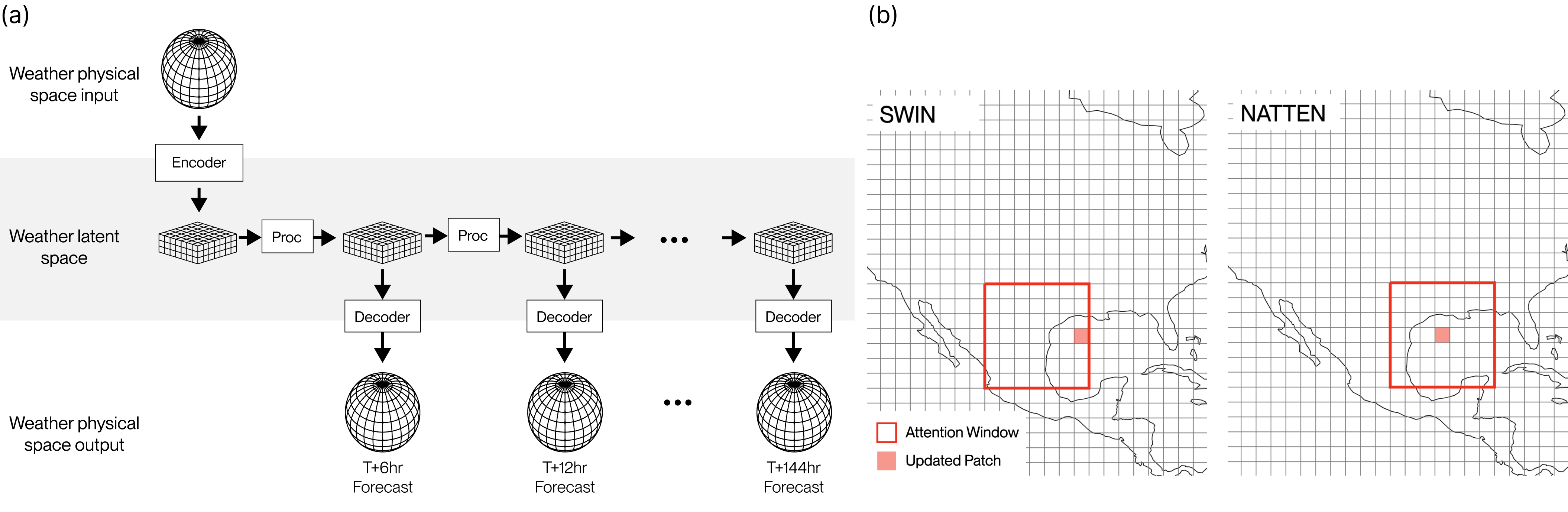}
    \caption{(a) A schematic of the encoder-processor-decoder architecture. (b) Illustration of the difference in attention window location between SWIN and NATTEN.}
    \label{fig:wm-highlevel}
\end{figure}

WM-3 consists of three high-level components: encoder, processor, and decoder.
The encoder has convolution layers that map the weather state from the 0.25-degree grid to a learned, lower-resolution (2-degree) latent space.
The processor then acts on the latent space any number of times, with each processor step corresponding to either one or six hours of forward time evolution.
Finally, a decoder with deconvolution layers translates the latent space back into a 3D grid in physical space.
A schematic of the model architecture is shown in Figure~\ref{fig:wm-highlevel}.

WM-3 uses a \emph{latent rollout} to produce a forecast for an arbitrary integer number of hours of forecast lead time in a single forward pass.
It achieves this by invoking a combination of the six-hour and one-hour processors repeatedly until the target lead time, using a simple greedy schedule. This avoids the typical MLWP rollout method of encoding a weather state, rolling forward six hours, decoding to the physical space, then re-encoding to take another six hour step forward. 
Keeping the atmospheric state in latent space between rollouts has three main advantages:
1) avoids extraneous encoding and decoding compute or error accumulation processes;
2) enables mixed-horizon prediction tasks in training (see Section~\ref{sec:training});
3) allows modularity in every model component by working with a shared latent space.

\paragraph{NATTEN}

The backbone of the WM-3 processors is a series of neighborhood attention-based transformer (NATTEN) blocks~\citep{hassani2023neighborhood}. 
In earlier versions of WeatherMesh, we experimented with the SWIN architecture~\citep{liu2021swin}, and found that NATTEN significantly improves WM-3 performance. 
The neighborhood attention mechanism provides a better inductive bias to the model for learning atmospheric physics due to its consistent locality of attention for information transfer between patches (Figure~\ref{fig:wm-highlevel}b).
In addition, thanks to the performance of the NATTEN library with Fused Neighborhood Attention kernels~\citep{hassani2024faster}, NATTEN is faster and has a lower memory footprint than our prior SWIN implementations.

\subsection{Training}\label{sec:training}

\paragraph{Pre-training} WeatherMesh-3 undergoes a two-stage training procedure: pretraining on ERA-5 reanalysis data, then fine-tuning on analysis data to create an operational version for real-time use. 

WM-3 is pretrained on ERA-5, ECMWF's reanalysis dataset, from 1979 to 2022, with 2020 left out of the training data for testing per \citet{rasp2024weatherbench}.
The six-hour processor is trained first to predict the state 6 and 12 hours later. The target timestep is then lengthened up to 120 hours (5 days, or 20 calls to the six-hour processor) according to a schedule.
Once the six-hour processor is trained, we freeze the weights in the encoder and decoder and train a one-hour processor.
We use Distributed Shampoo~\citep{shi2023distshampoo} for training on multiple GPUs.
See Appendix~\ref{app:training} for details on the training setup and \ref{app:matepoint} for compute optimizations.

\paragraph{Operationalization} 
In order to make WM-3 suitable for operational use, we replace the pretrained encoder with two new encoders that ingest analyses from European Centre for Medium-Range Weather Forecasts (ECMWF)'s Integrated Forecasting System (IFS) and the National Oceanic and Atmospheric Administration (NOAA)'s Global Forecast System (GFS). See Figure~\ref{fig:wm-realtime} for details.
WeatherMesh has been running operationally since March 2024.

\section{Results}\label{sec:results}

\paragraph{Accuracy}
Operational WM-3 achieves high accuracy relative to current operational models.
To evaluate WM-3, we calculate latitude-weighted RMSE scores for our operational forecasts against ERA-5 and compare with IFS HRES over our evaluation period in 2024 (see Appendix~\ref{app:validation-details}).
In Figure~\ref{fig:scorecard-wm3-hres}, we present the evaluation scorecards of WM-3. 
WM-3 outperforms HRES on \emph{all but one} of 690 targets evaluated (geopotential at 50 hPa at 10 days), with a notable 37.7\% improvement over HRES on 2-meter temperature at 1 day lead time. 

\begin{figure}[h!]
    \centering
    \includegraphics[width=\linewidth]{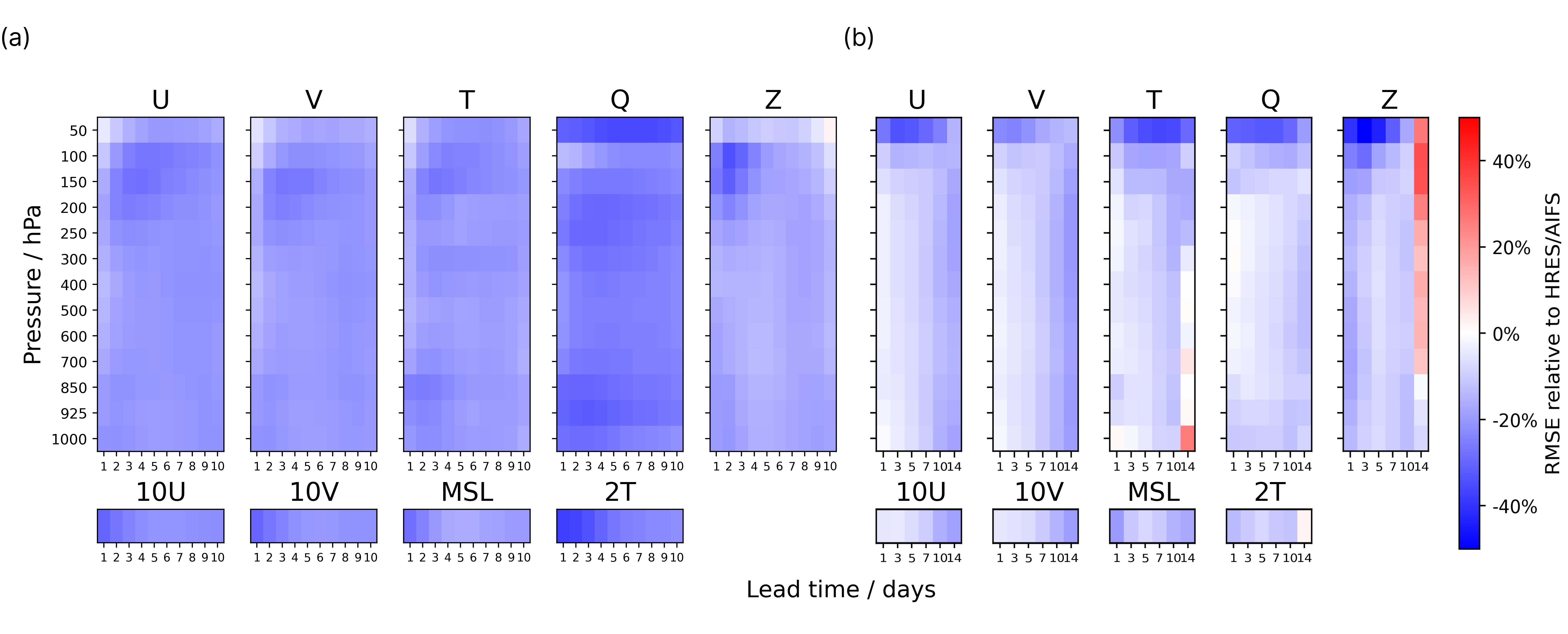}
    \caption{(a) Scorecard comparing WM-3 vs. IFS HRES across all pressure levels and surface variables. (b) Scorecard comparing WM-3 vs. AIFS.}
    \label{fig:scorecard-wm3-hres}
\end{figure}

MLWP models outperforming HRES on RMSE should be evaluated with the perspective that HRES has not been optimized for RMSE as a metric. While it is common to compute RMSE for HRES against its own analysis in literature~\citep{lam2023graphcast, bodnar2024aurora}, we report RMSE against ERA-5 for both models and treat ERA-5 as the best representation of ground truth.

We also report WM-3's performance against AIFS, ECMWF's own operational AI model~\citep{lang2024aifs}, in Figure~\ref{fig:scorecard-wm3-hres}(b). We note that WM-3 consistently outperforms AIFS at 50 hPa but is relatively weaker at 14 days forecast lead time, especially on geopotential (Z).

\paragraph{Blurring}

\begin{figure}[h!]
    \centering
    \includegraphics[width=\linewidth]{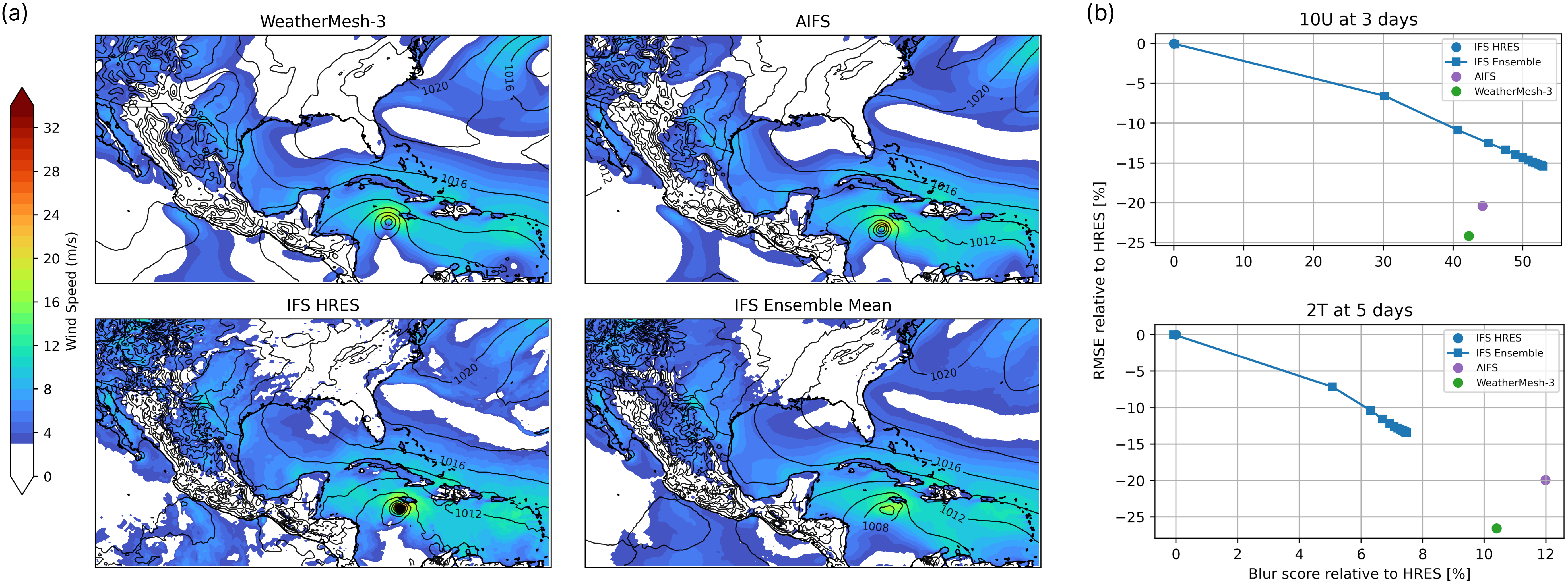}
    \caption{(a) Forecast maps showing 10-meter windspeed and mean sea level pressure at 72 hours lead time from 00z on July, 1, 2024. MLWP models offer notably less fine-grained spatial details, but resolve the intensity of Hurricane Beryl more clearly than the IFS ensemble mean. (b) Blur score vs RMSE for select surface variables. The points on the IFS Ensemble line represent ensemble means for member subsets beginning at 1 member and ending at 51 members.}
    \label{fig:blur}
\end{figure}

MLWP models are known to suffer from excessive blurring~\citep{lam2023graphcast, rasp2024weatherbench, lang2024aifscrps}, where models learn to blur excessively to achieve lower training loss (usually mean squared error, MSE).
In operational use, however, a blurry forecast with a lower RMSE score is often less useful than a less blurry forecast with a slightly higher RMSE score. 
To evaluate WM-3 in the trade-off between accuracy and blur, we calculate a ``blur score'' based on spectral power at 500km wavelength (defined in Appendix~\ref{app:validation-details}), and plot it against RMSE, shown in Figure~\ref{fig:blur}(b).
A useful reference is ECMWF's ensemble forecasts (ENS)~\citep{ifsens}, which leverage an ensemble of 51 members to achieve higher accuracy but inevitably introduces more blurring. 
WM-3 simultaneously achieves lower RMSE score \emph{and} lower blur score than both ENS and AIFS on some variables and forecast lead times.
Furthermore, for most variables and lead times, WM-3 attains more accuracy gain for a given level of blurring (or less blurring for a given level of accuracy) than both ENS and AIFS.
We present these plots for a wider range of variables and lead times in Appendix~\ref{app:additional-results}.

\section{Discussion}\label{sec:discussion}

\subsection{Forecast Accessibility}

\paragraph{Compute efficiency}
The compute efficiency of WM-3 represents a significant step toward democratizing access to high-quality weather forecasts. By reducing hardware requirements by several orders of magnitude compared to traditional NWP systems while maintaining or exceeding their accuracy, WM-3 enables organizations with limited computational resources to run their own customized forecasting systems. A 14-day global forecast can be computed in just 12 seconds on a single RTX 4090, roughly 143,000 times faster (in node seconds per hour lead time) than physics-based numerical weather models~\citep{bodnar2024aurora}. On a single H100 server, hourly global forecasts out to 14 days can be computed in under 10 seconds.

WM-3 runs on hardware as light as a consumer grade laptop with 16GB VRAM and 32GB RAM, which is one third of Aurora's VRAM usage~\citep{bodnar2024aurora}. This accessibility has the potential to benefit entities that previously lacked access to advanced forecasting capabilities due to infrastructure constraints, democratizing access to climate resilience tools.

\paragraph{Modular customization}
The architecture of WM-3 enables users to ingest additional operational data sources for modular customization. Operational WM-3 currently ingests ECMWF IFS and NOAA GFS analyses via two separate encoders, trained as described in Section \ref{sec:training}. The architecture is well-suited for users to create modular extensions that ingest additional custom data sources to enhance forecast accuracy. Accessibility to this level of forecast customization will become increasingly pertinent for users in light of rising need for forecasting weather and extreme events.





\subsection{Future Work}
Our results suggest several directions for advancing this research. In particular, the computational efficiency of WM-3 makes large ensemble forecasts feasible on modest hardware, opening new possibilities for novel ensemble techniques. These could leverage the model's speed to generate more sophisticated uncertainty quantification and improve forecast reliability.
This computational advantage also enables fast integration of additional data sources to improve operational forecasts via a live data assimilation pipeline. This real-time data can be incorporated through additional encoder pathways while maintaining the efficient latent space processing that enables our current performance gains. We are eager to continue future work towards operationalizing fast, accurate, and accessible forecasts. 


\clearpage
\bibliography{iclr2025_conference}

\begin{thebibliography}{18}
\providecommand{\natexlab}[1]{#1}
\providecommand{\url}[1]{\texttt{#1}}
\expandafter\ifx\csname urlstyle\endcsname\relax
  \providecommand{\doi}[1]{doi: #1}\else
  \providecommand{\doi}{doi: \begingroup \urlstyle{rm}\Url}\fi

\bibitem[Bi et~al.(2023)Bi, Xie, Zhang, Chen, Gu, and Tian]{bi2023pangu}
Kaifeng Bi, Lingxi Xie, Hengheng Zhang, Xin Chen, Xiaotao Gu, and Qi~Tian.
\newblock Accurate medium-range global weather forecasting with 3d neural networks.
\newblock \emph{Nature}, 619\penalty0 (7970):\penalty0 533--538, 2023.

\bibitem[Bodnar et~al.(2024)Bodnar, Bruinsma, Lucic, Stanley, Brandstetter, Garvan, Riechert, Weyn, Dong, Vaughan, et~al.]{bodnar2024aurora}
Cristian Bodnar, Wessel~P Bruinsma, Ana Lucic, Megan Stanley, Johannes Brandstetter, Patrick Garvan, Maik Riechert, Jonathan Weyn, Haiyu Dong, Anna Vaughan, et~al.
\newblock Aurora: A foundation model of the atmosphere.
\newblock \emph{arXiv preprint arXiv:2405.13063}, 2024.

\bibitem[Chen et~al.(2023)Chen, Zhong, Zhang, Cheng, Xu, Qi, and Li]{chen2023fuxi}
Lei Chen, Xiaohui Zhong, Feng Zhang, Yuan Cheng, Yinghui Xu, Yuan Qi, and Hao Li.
\newblock Fuxi: A cascade machine learning forecasting system for 15-day global weather forecast.
\newblock \emph{npj Climate and Atmospheric Science}, 6\penalty0 (1):\penalty0 190, 2023.

\bibitem[Dosovitskiy(2020)]{dosovitskiy2020vit}
Alexey Dosovitskiy.
\newblock An image is worth 16x16 words: Transformers for image recognition at scale.
\newblock \emph{arXiv preprint arXiv:2010.11929}, 2020.

\bibitem[ECMWF(2024)]{ifsens}
ECMWF.
\newblock \emph{IFS Documentation CY49R1 - Part V: Ensemble Prediction System}, chapter~5.
\newblock ECMWF, 11/2024 2024.
\newblock \doi{10.21957/956d60ad81}.

\bibitem[Gupta et~al.(2018)Gupta, Koren, and Singer]{gupta2018shampoo}
Vineet Gupta, Tomer Koren, and Yoram Singer.
\newblock Shampoo: Preconditioned stochastic tensor optimization.
\newblock In \emph{International Conference on Machine Learning}, pp.\  1842--1850. PMLR, 2018.

\bibitem[Hassani et~al.(2023)Hassani, Walton, Li, Li, and Shi]{hassani2023neighborhood}
Ali Hassani, Steven Walton, Jiachen Li, Shen Li, and Humphrey Shi.
\newblock Neighborhood attention transformer.
\newblock In \emph{IEEE/CVF Conference on Computer Vision and Pattern Recognition (CVPR)}, 2023.

\bibitem[Hassani et~al.(2024)Hassani, Hwu, and Shi]{hassani2024faster}
Ali Hassani, Wen-Mei Hwu, and Humphrey Shi.
\newblock Faster neighborhood attention: Reducing the $o(n^2)$ cost of self attention at the threadblock level.
\newblock In \emph{Advances in Neural Information Processing Systems}, 2024.

\bibitem[He et~al.(2016)He, Zhang, Ren, and Sun]{he2016resnet}
Kaiming He, Xiangyu Zhang, Shaoqing Ren, and Jian Sun.
\newblock Deep residual learning for image recognition.
\newblock In \emph{Proceedings of the IEEE conference on computer vision and pattern recognition}, pp.\  770--778, 2016.

\bibitem[Kurth et~al.(2023)Kurth, Subramanian, Harrington, Pathak, Mardani, Hall, Miele, Kashinath, and Anandkumar]{kurth2023fourcastnet}
Thorsten Kurth, Shashank Subramanian, Peter Harrington, Jaideep Pathak, Morteza Mardani, David Hall, Andrea Miele, Karthik Kashinath, and Anima Anandkumar.
\newblock Fourcastnet: Accelerating global high-resolution weather forecasting using adaptive fourier neural operators.
\newblock In \emph{Proceedings of the platform for advanced scientific computing conference}, pp.\  1--11, 2023.

\bibitem[Lam et~al.(2023)Lam, Sanchez-Gonzalez, Willson, Wirnsberger, Fortunato, Alet, Ravuri, Ewalds, Eaton-Rosen, Hu, et~al.]{lam2023graphcast}
Remi Lam, Alvaro Sanchez-Gonzalez, Matthew Willson, Peter Wirnsberger, Meire Fortunato, Ferran Alet, Suman Ravuri, Timo Ewalds, Zach Eaton-Rosen, Weihua Hu, et~al.
\newblock Learning skillful medium-range global weather forecasting.
\newblock \emph{Science}, 382\penalty0 (6677):\penalty0 1416--1421, 2023.

\bibitem[Lang et~al.(2024{\natexlab{a}})Lang, Alexe, Chantry, Dramsch, Pinault, Raoult, Clare, Lessig, Maier-Gerber, Magnusson, et~al.]{lang2024aifs}
Simon Lang, Mihai Alexe, Matthew Chantry, Jesper Dramsch, Florian Pinault, Baudouin Raoult, Mariana~CA Clare, Christian Lessig, Michael Maier-Gerber, Linus Magnusson, et~al.
\newblock Aifs-ecmwf's data-driven forecasting system.
\newblock \emph{arXiv preprint arXiv:2406.01465}, 2024{\natexlab{a}}.

\bibitem[Lang et~al.(2024{\natexlab{b}})Lang, Alexe, Clare, Roberts, Adewoyin, Bouall{\`e}gue, Chantry, Dramsch, Dueben, Hahner, et~al.]{lang2024aifscrps}
Simon Lang, Mihai Alexe, Mariana~CA Clare, Christopher Roberts, Rilwan Adewoyin, Zied~Ben Bouall{\`e}gue, Matthew Chantry, Jesper Dramsch, Peter~D Dueben, Sara Hahner, et~al.
\newblock Aifs-crps: Ensemble forecasting using a model trained with a loss function based on the continuous ranked probability score.
\newblock \emph{arXiv preprint arXiv:2412.15832}, 2024{\natexlab{b}}.

\bibitem[Liu et~al.(2021)Liu, Lin, Cao, Hu, Wei, Zhang, Lin, and Guo]{liu2021swin}
Ze~Liu, Yutong Lin, Yue Cao, Han Hu, Yixuan Wei, Zheng Zhang, Stephen Lin, and Baining Guo.
\newblock Swin transformer: Hierarchical vision transformer using shifted windows.
\newblock In \emph{Proceedings of the IEEE/CVF international conference on computer vision}, pp.\  10012--10022, 2021.

\bibitem[Price et~al.(2023)Price, Sanchez-Gonzalez, Alet, Andersson, El-Kadi, Masters, Ewalds, Stott, Mohamed, Battaglia, et~al.]{price2023gencast}
Ilan Price, Alvaro Sanchez-Gonzalez, Ferran Alet, Tom~R Andersson, Andrew El-Kadi, Dominic Masters, Timo Ewalds, Jacklynn Stott, Shakir Mohamed, Peter Battaglia, et~al.
\newblock Gencast: Diffusion-based ensemble forecasting for medium-range weather.
\newblock \emph{arXiv preprint arXiv:2312.15796}, 2023.

\bibitem[Rasp et~al.(2024)Rasp, Hoyer, Merose, Langmore, Battaglia, Russell, Sanchez-Gonzalez, Yang, Carver, Agrawal, et~al.]{rasp2024weatherbench}
Stephan Rasp, Stephan Hoyer, Alexander Merose, Ian Langmore, Peter Battaglia, Tyler Russell, Alvaro Sanchez-Gonzalez, Vivian Yang, Rob Carver, Shreya Agrawal, et~al.
\newblock Weatherbench 2: A benchmark for the next generation of data-driven global weather models.
\newblock \emph{Journal of Advances in Modeling Earth Systems}, 16\penalty0 (6):\penalty0 e2023MS004019, 2024.

\bibitem[Shi et~al.(2023)Shi, Lee, Iwasaki, Gallego-Posada, Li, Rangadurai, Mudigere, and Rabbat]{shi2023distshampoo}
Hao-Jun~Michael Shi, Tsung-Hsien Lee, Shintaro Iwasaki, Jose Gallego-Posada, Zhijing Li, Kaushik Rangadurai, Dheevatsa Mudigere, and Michael Rabbat.
\newblock A distributed data-parallel pytorch implementation of the distributed shampoo optimizer for training neural networks at-scale.
\newblock \emph{arXiv preprint arXiv:2309.06497}, 2023.

\bibitem[Su et~al.(2024)Su, Ahmed, Lu, Pan, Bo, and Liu]{su2024rope}
Jianlin Su, Murtadha Ahmed, Yu~Lu, Shengfeng Pan, Wen Bo, and Yunfeng Liu.
\newblock Roformer: Enhanced transformer with rotary position embedding.
\newblock \emph{Neurocomputing}, 568:\penalty0 127063, 2024.

\end{thebibliography}
\bibliographystyle{iclr2025_conference}
\clearpage
\appendix
\section{Model Details}\label{sec:app}

\subsection{Variables}

WM-3 forecasts 17 surface variables and 5 atmospheric variables. Each variable is presented in Table~\ref{tab:vars} below with its corresponding ID from the \href{https://codes.ecmwf.int/grib/param-db/}{ECMWF Parameter Database}, which contains further information. All atmospheric variables are forecasted at 28 pressure levels: 10, 30, 50, 70, 100, 125, 150, 175, 200, 250, 300, 350, 400, 450, 500, 550, 600, 650, 700, 750, 800, 850, 875, 900, 925, 950, 975, 1000 hPa. 

\begin{table}[h]
    \centering
    \setlength{\extrarowheight}{3pt}
    \begin{tabular}{|l|c|}
        \hline
        \textbf{Variable Name} & \textbf{ECMWF Parameter ID} \\
        \hline
        \multicolumn{2}{|c|}{\textbf{Surface Variables}} \\
        \hline
        10 metre u wind component (10U) & 165 \\
        10 metre v wind component (10V) & 166 \\
        2 metre temperature (2T) & 167 \\
        Mean sea-level pressure (MSL) & 151 \\
        Total cloud cover & 45 \\
        2 metre dewpoint temperature & 168 \\
        100 metre u wind component & 246 \\
        100 metre v wind component & 247 \\
        Mean shortwave radiation flux & 15 \\
        Large-scale precipitation & 142 \\
        Convective precipitation & 143 \\
        Maximum 2 metre temperature & 201 \\
        Minimum 2 metre temperature & 202 \\
        Large-scale precipitation (6h) & 142 \\
        Convective precipitation (6h) & 143 \\
        Maximum 2 metre temperature (6h) & 201 \\
        Minimum 2 metre temperature (6h) & 202 \\
        \hline
        \multicolumn{2}{|c|}{\textbf{Atmospheric Variables}} \\
        \hline
        Geopotential (Z) & 129 \\
        Temperature (T) & 130 \\
        U component of wind (U) & 131 \\
        V component of wind (V) & 132 \\
        Specific humidity (Q) & 133 \\
        \hline
    \end{tabular}
    \caption{ECMWF Parameter IDs for WM-3 surface and atmospheric variables.}
    \label{tab:vars}
\end{table}

Most of the surface variables are not available in IFS or GFS analyses, and WM-3 takes as input only the first eight surface variables.

At 0.25 resolution, input and output are arrays of 721 $\times$ 1440. In order to optimize for storage space as well as computational efficiency, we omit the last latitude row (the south pole) and only operate with arrays of 720 $\times$ 1440.

\subsection{Model Architecture}\label{app:arch}

The encoder consists of ResNet blocks~\citep{he2016resnet} and convolutional layers. 
The following constant variables are concatenated with the input: sine and cosine of latitudes and longitutes, the sea-land mask, soil type, topography, and elevation. 
The encoder includes 2 NATTEN blocks with window size (5, 7, 7) in depth, width, height, corresponding to the vertical height, latitude, and longitude dimensions respectively. The hidden dimension is 1024.

Both processors consist of 10 NATTEN blocks.

The decoder is the reverse of the encoder: it begins with 2 NATTEN blocks, and then a series of ResNet and deconvolutional layers take the output from latent space back to physical space.

To make NATTEN work on a sphere, we implement our own circular padding. 
At the poles, we use the bump attention behavior from NATTEN. 
For position encoding of tokens, we use rotary embeddings~\citep{su2024rope}.

We share our model code in a \href{https://github.com/windborne/WeatherMesh-3}{repository} that we open source as part of this submission.

\subsection{Training details}\label{app:training}

WeatherMesh-3 is pretrained on ERA-5 data from January 23, 1979 to December 28, 2019, as well from February 1, 2021 to December 21, 2023. The year of 2020 is left out of the training set for evaluation as is conventional~\citep{rasp2024weatherbench}. 
It is trained for 42,000 steps on a cosine learning rate schedule with a maximum learning rate of 3e-4.
The training target is mean squared error (MSE) at target timestep dt, with dt determined by a schedule. The schedule is as follows for the 6-hour processor:
\begin{table}[h!]
    \centering
    \setlength{\extrarowheight}{2pt}
    \begin{tabular}{|l|l|}
        \hline
        \textbf{Starting step} & \textbf{Target dt hours} \\
        \hline
        0 & 0, 6, 12 \\
        1,000 & 0, 6, 12, 18, 24 \\
        15,000 & 0, 6, 12, 18, 24, 30 \\
        21,000 & 0, 6, 12, 18, 24, 30, 36 \\
        26,000 & 0, 6, 12, 18, 24, 30, 36, 42 \\
        30,000 & 0, 6, 12, 18, 24, 30, 36, 42, 48 \\
        \hline
    \end{tabular}
    \caption{Schedule for introducing additional target dts during training.}
    \label{tab:schedule}
\end{table}

At each step, a random subset of 5 dts are chosen as training target, while always including the largest dt allowed to make sure the time it takes to step across GPUs is the same.

The model then undergoes another annealing cycle of 16,000 steps training on dts up to 120 hours (5 days).
The 1-hour processor, trained with a frozen encoder and decoder, is trained for 25,000 steps on 5 random dts between 0 and 24 at each step.

\begin{figure}[h!]
    \centering
    \includegraphics[width=0.75\linewidth]{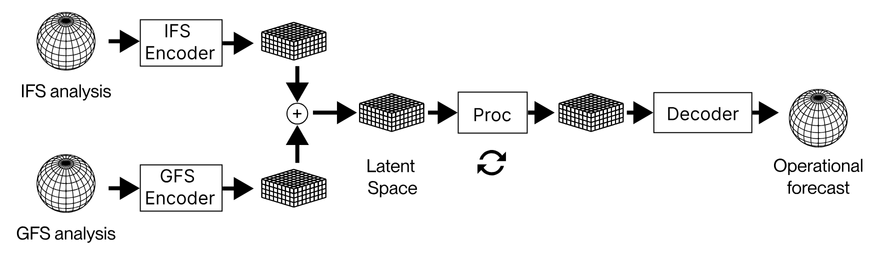}
    \caption{Schematic for operational encoders.}
    \label{fig:wm-realtime}
\end{figure}

For operational fine-tuning, we attach two new encoders to take IFS and GFS analyses as input, as shown in Figure~\ref{fig:wm-realtime}. We train on IFS and GFS analyses from March 2021 to February 2024.

We use the PyTorch implementation of distributed shampoo~\citep{shi2023distshampoo, gupta2018shampoo}, a second-order optimizer, and train across 6 RTX 4090 GPUs. 
We use half precision in training.

\subsection{Validation details}\label{app:validation-details}

For results in Figure~\ref{fig:scorecard-wm3-hres}, we used a validation period of approximately five months, from March 10, 2024 to July 31, 2024, and evaluated forecasts initialized at 0z. For results in Figure~\ref{fig:blur}, the validation period is March 21, 2024 to July 31, 2024, again with all forecasts initialized at 0z. The dates are chosen due to the availability of data at the time of writing, not as a result of cherry-picking. 

For the accuracy metric, we calculate latitutde-weighted root mean squared error (RMSE)
\begin{equation}
    \text{RMSE} = \frac{1}{T} \sum_{t=1}^T \sqrt{\frac{1}{H\times W} \sum_{i=1}^H \sum_{j=1}^W w(i) (\hat{X}^t_{i,j} - X^t_{i,j})^2},
\end{equation}
where $\hat{X}$ is the predicted value, $X$ is the ground truth value, $t$ ranges over evaluation dates, $i$ and $j$ range over latitude and longitude ($H = 720$ and $W = 1440$ at 0.25 degrees), and $w(i) = \cos\left(\frac{ i \cdot \pi}{720}\right)$
is the latitude weighting factor.

For the blur metric, we calculate a blur score
\begin{equation}
    \text{Blur score} = \frac{1}{\sqrt{S_{500}}},
\end{equation}
where $S_{500}$ is the spectral power at a wavelength of 500 km, as defined in \citet{rasp2024weatherbench}.
The wavelength of 500 km is somewhat arbitrary, but is meant to capture blurring on the scale of e.g. the size of a hurricane without being dominated by smaller artifacts present in a 0.25-degree (25-km) resolution forecast.

\subsection{Matepoint}\label{app:matepoint}
A single training sample for weather forecasting requires a lot of computation: one cannot predict weather for only half of the globe at a time, the entire globe must be forecasted at once as it is an interconnected system. Weather over the entire earth is a much larger input than a passage of text in the context window of an LLM or an image. For this reason, we currently train with only a batch size of 1.

It is not possible to train a global weather model without careful considerations of VRAM utilization. Storing the intermediate model activations for a backwards pass for even the shortest forecast step immediately takes hundreds of GiB of VRAM if done naively. As is common practice, we opt to do model checkpoints for the main transformer blocks to dramatically save VRAM at the cost of longer backwards pass compute time. This works, but it breaks down for training longer forecast time horizons.

The challenge with conventional checkpointing is that it requires storing the input to the checkpointed layer. For a transformer, this means storing the latent space for each transformer block. A 6 day forecast requires running over 200 transformer layers, so for a latent space of 200MiB, this quickly takes up 40GiB of VRAM just for these tensors alone. WM-3 was trained entirely on RTX 4090s due to their very attractive cost per compute, and so this would be entirely unworkable.

The solution is an additional step over just checkpointing. Rather than keep the input tensor to a checkpoint on the GPU, we send it back to the CPU to be stored in RAM, a much more abundant resource. Because forecasting longer in time simply means calling more checkpointed transformer layers, this method means that there is zero VRAM cost to longer forecasting during training time, and there is effectively no limit.

We create a forked version of the pytorch checkpoint library named ``matepoint'' to implement this concept. There are a few computational considerations required in order to do this without slowing down training. First, tensors must be sent back to the GPU on an independent CUDA stream to allow for full parallelization. Second, tensors that are needed during the backwards pass must be pipelined so that they arrive on the GPU before they are needed. If tensors were only moved as needed naively, then this would incur a delay between each transformer layer in the backwards pass while waiting on tensors to arrive. 
We have also released the library as an open source python \href{https://github.com/windborne/matepoint}{package}.




\subsection{Additional results}\label{app:additional-results}

\begin{figure}
    \centering
    \includegraphics[width=\linewidth]{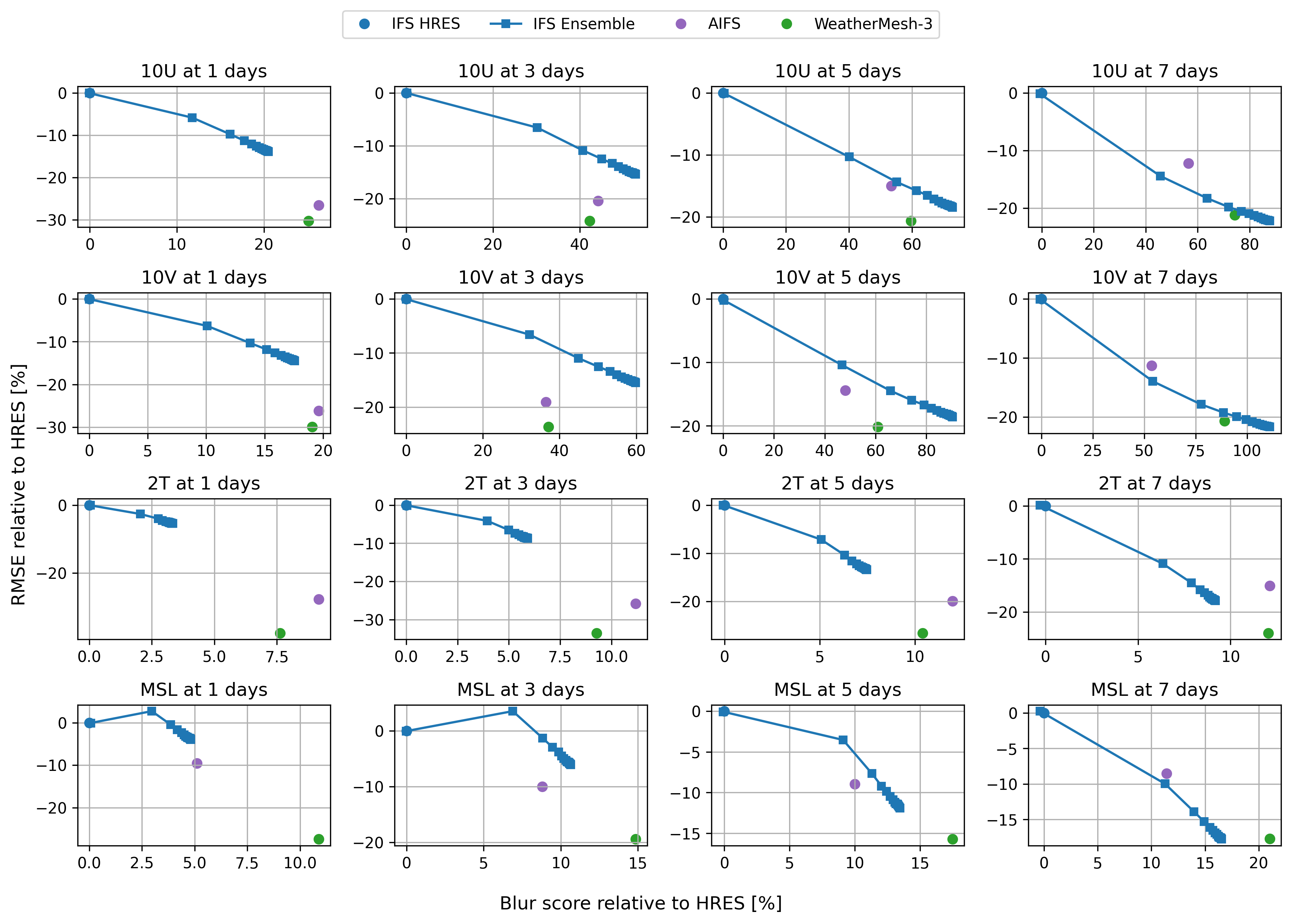}
    \caption{RMSE-blur score for surface variables.}
    \label{app:fig:blur-sfc}
\end{figure}

\begin{figure}
    \centering
    \includegraphics[width=\linewidth]{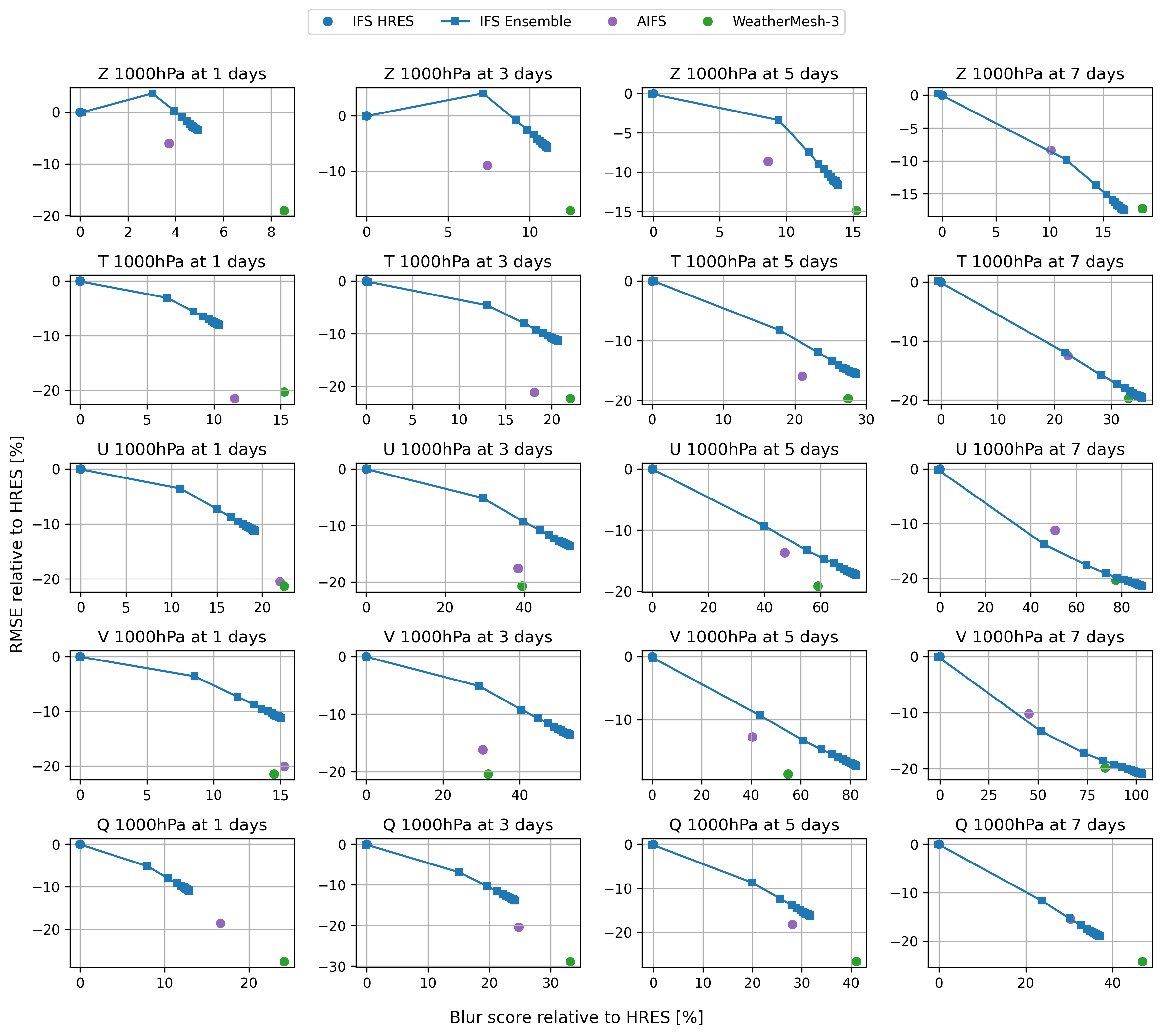}
    \caption{RMSE-blur score for variables at 1000 hPa.}
    \label{app:fig:blur-1000}
\end{figure}

\begin{figure}
    \centering
    \includegraphics[width=\linewidth]{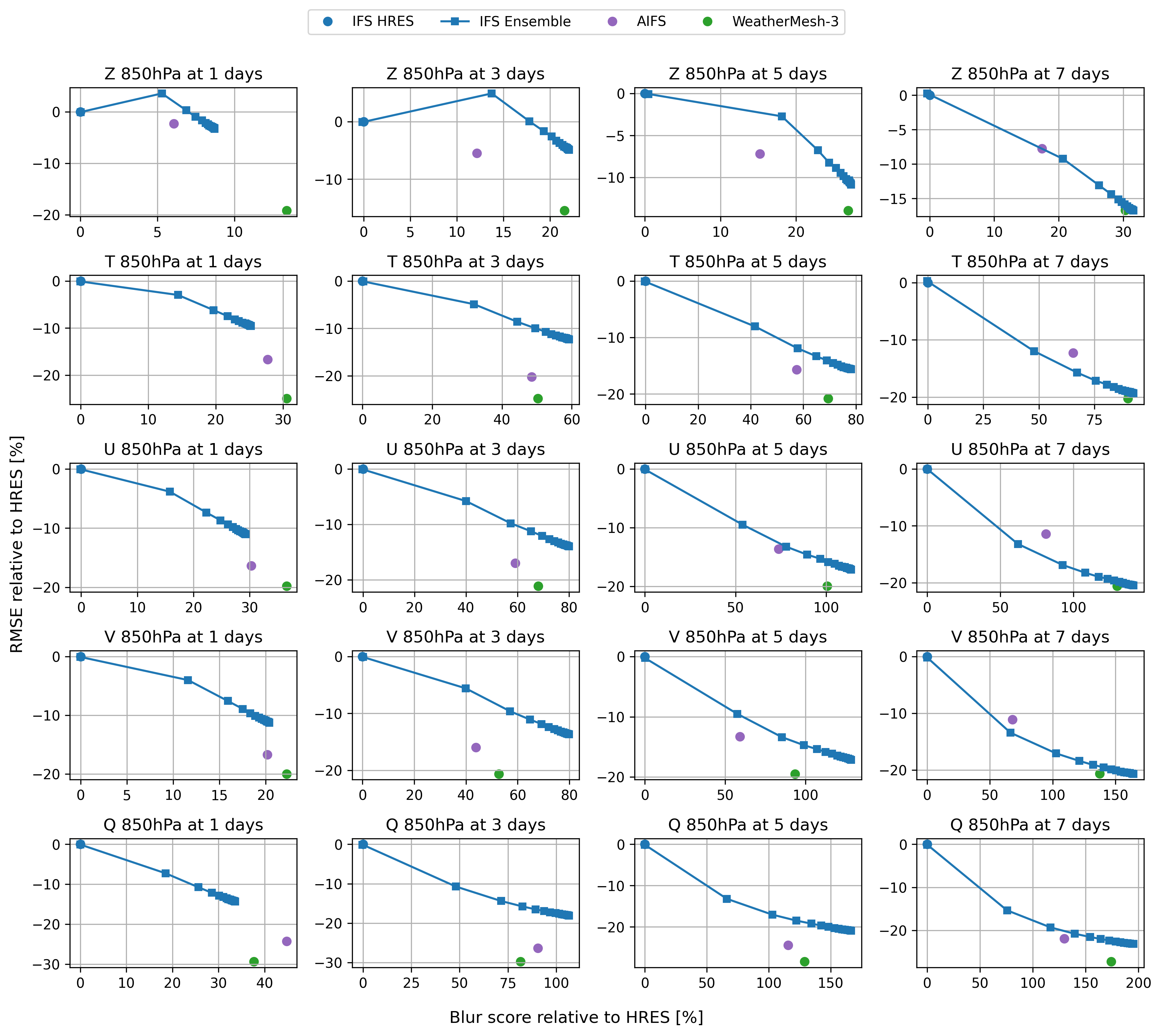}
    \caption{RMSE-blur score for variables at 850 hPa.}
    \label{app:fig:blur-850}
\end{figure}

\begin{figure}
    \centering
    \includegraphics[width=\linewidth]{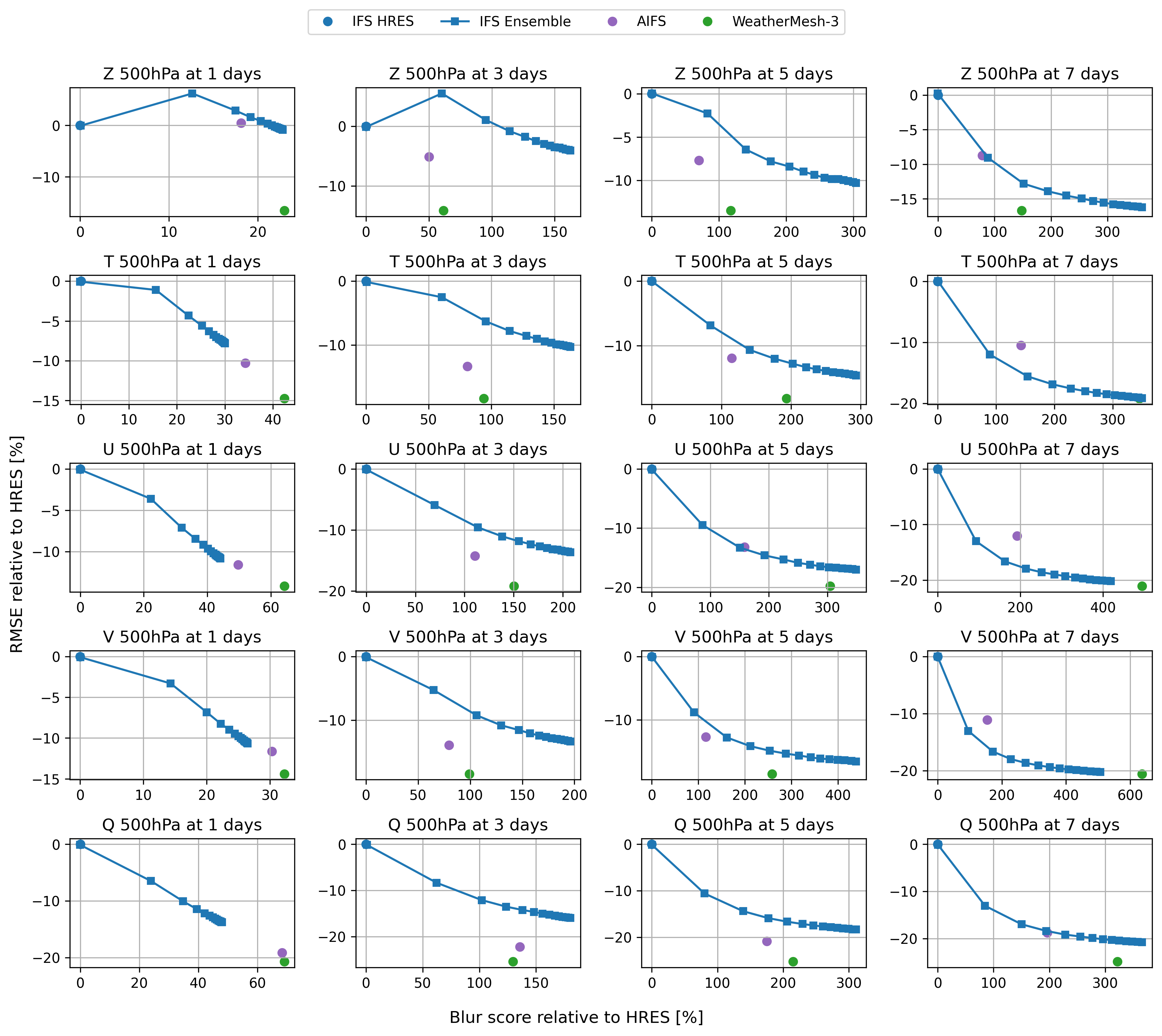}
    \caption{RMSE-blur score for variables at 500 hPa.}
    \label{app:fig:blur-500}
\end{figure}

\begin{figure}
    \centering
    \includegraphics[width=\linewidth]{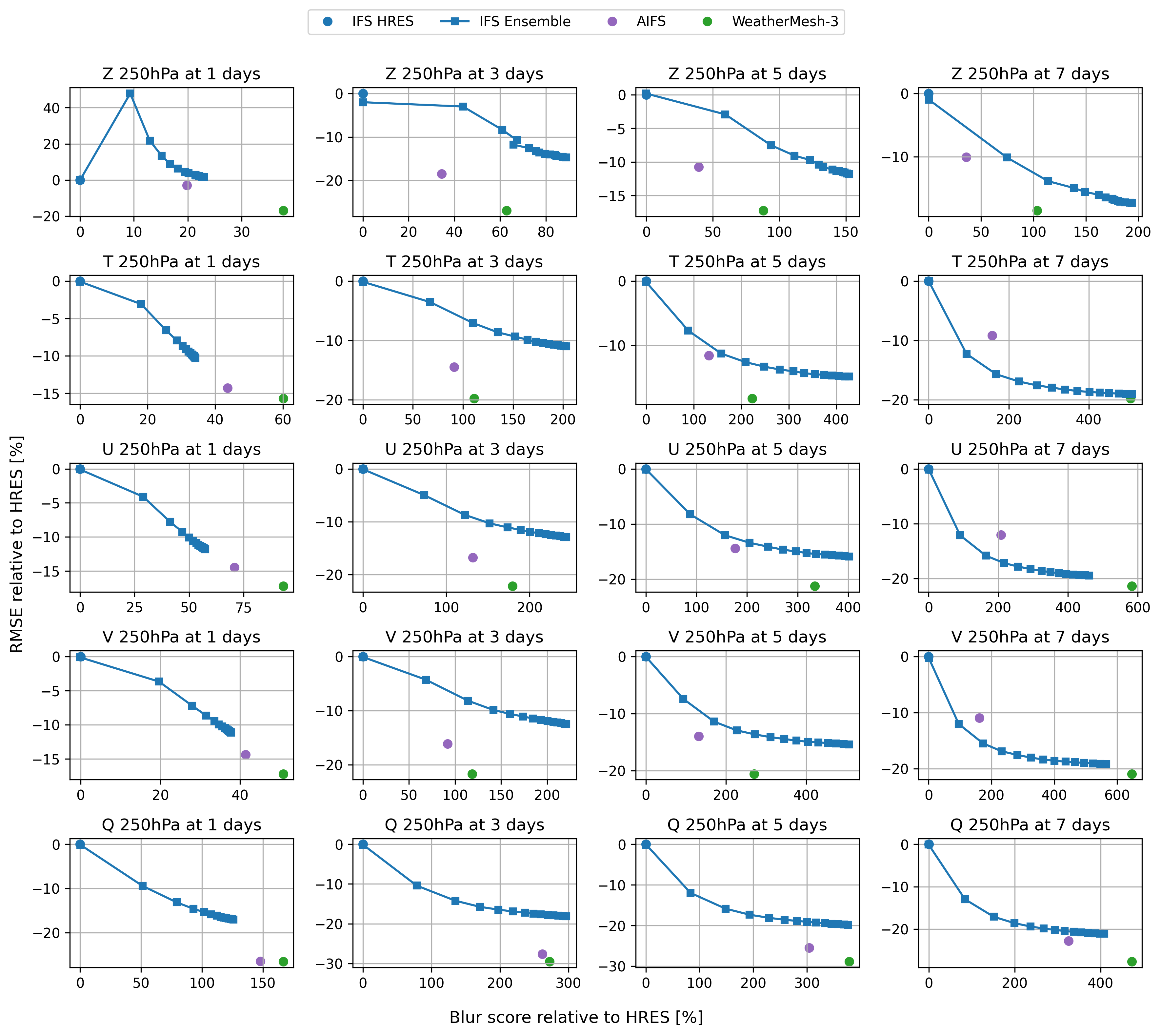}
    \caption{RMSE-blur score for variables at 250 hPa.}
    \label{app:fig:blur-250}
\end{figure}

We present the RMSE-blur score plots for a wider range of variables in Figures~\ref{app:fig:blur-sfc}--\ref{app:fig:blur-250}. The number of IFS ensemble members included to produce the IFS Ensemble line in these RMSE-blur score plots is 1, 2, 4, 8, 12, 16, 20, 24, 28, 32, 36, 40, 44, 48, and 51.

\end{document}